\pdfoutput=1
\documentclass[10pt,twocolumn,letterpaper]{article}

\usepackage[pagenumbers]{wacv}

\usepackage{booktabs}
\usepackage{colortbl}
\definecolor{rlbestblue}{RGB}{0,83,170}
\definecolor{bestred}{RGB}{220,0,0}
\definecolor{rlbestgreen}{RGB}{0,104,72}
\definecolor{oursrowgray}{RGB}{228,228,228}

\usepackage{graphicx}
\graphicspath{{figs/}}
\usepackage{tikz}
\usetikzlibrary{calc}
\usepackage{pgfplots}
\pgfplotsset{compat=1.18}
\usepgfplotslibrary{groupplots}

\definecolor{wacvblue}{rgb}{0.21,0.49,0.74}
\usepackage[breaklinks,colorlinks,allcolors=wacvblue]{hyperref}
\hypersetup{
    pdftitle={MR-IQA: A Unified Margin View of Regression and Ranking for Blind Image Quality Assessment},
    pdfauthor={Yuan Li, Youyuan Lin, Zitang Sun, Yung-Hao Yang, Kiyofumi Miyoshi, Chenhui Chu, Shin'ya Nishida}
}
\makeatletter
\providecommand\protected@file@percent{}
\providecommand{\@LN}[2]{}
\makeatother
\newcommand{\arxivrighttopfig}{}
\newcommand{\arxivCodeAvailability}{Code is available at \href{https://github.com/RobinY99/MR-IQA.git}{MR-IQA}.}

\newcommand{\bridgeRegressionRankingTitle}{\texorpdfstring{\bfseries\normalsize Bridge Between Regression and Ranking}{Bridge Between Regression and Ranking}}

\title{MR-IQA: A Unified Margin View of Regression and Ranking for Blind Image Quality Assessment}

\author{
{\normalsize Yuan Li\enspace Youyuan Lin\enspace Zitang Sun\enspace Yung-Hao Yang\enspace Kiyofumi Miyoshi\enspace Chenhui Chu\enspace Shin'ya Nishida}\\
{\normalsize Graduate School of Informatics, Kyoto University}
}
\date{}

\begin{document}
\maketitle
\begin{abstract}
Blind image quality assessment (BIQA) is commonly built on two basic learning paradigms: regression and ranking. Regression calibrates absolute scores, whereas ranking recovers quality structure from ordinal relations. Although joint regression-ranking supervision often improves BIQA, the relation between the two paradigms remains largely empirical and underexplored. In this work, we revisit what underlies regression and ranking and identify pairwise relational distance, termed quality margin, as their common bridge. Our derivation shows that, at the objective-optimization level, both paradigms fit quality margins: regression fits margins induced by score endpoints, while ranking fits transformed or sign-level margins through preference probabilities. Motivated by this insight, we propose MR-IQA, a direct quality-margin optimization framework for reinforcement learning (RL)-based BIQA. MR-IQA samples quality scores and optimizes pairwise margin errors as policy rewards, thereby modeling quality structure more explicitly. Experiments on six BIQA benchmarks show competitive general performance, and controlled comparisons demonstrate that MR-IQA achieves the strongest average PLCC/SRCC over regression- or ranking-based RL methods. Our findings provide a new insight into unifying regression and ranking, offering a theoretical basis for understanding quality-structure modeling in BIQA and beyond. \ifdefined\arxivCodeAvailability \arxivCodeAvailability\fi
\end{abstract}

\section{Introduction}
\label{sec:intro}

Blind image quality assessment (BIQA) seeks to model how humans judge perceptual image quality from visual content. Along with the development of machine learning, BIQA has evolved from hand-crafted statistical priors~\cite{NIQE,BRISQUE} to deep visual representations~\cite{dbcnn,musiq,yang2022maniqa,wang2023exploring,arniqa,RALI} and, more recently, to multimodal large language models (MLLMs)~\cite{q-align,deqa,q-insight,depictqa,q-instruct}. These frameworks have expanded BIQA from score prediction alone toward a unified assessment interface that can produce language-based quality reasoning.

\ifdefined\arxivrighttopfig
\newpage
\noindent\begin{minipage}{\columnwidth}
\setlength{\abovecaptionskip}{2pt}
\setlength{\belowcaptionskip}{0pt}
\centering
\includegraphics[width=\columnwidth]{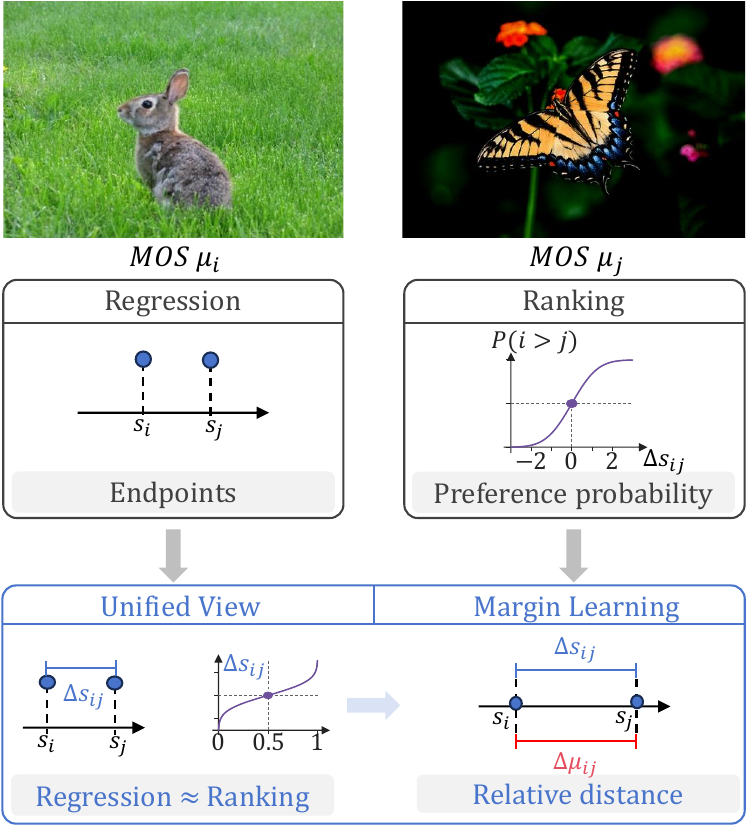}
\captionsetup{type=figure,hypcap=false}
\caption{Motivation of margin learning for BIQA. Regression estimates pointwise quality endpoints, while ranking maps pairwise differences into preference probabilities. Both methods can be interpreted through a unified margin view, as formalized in \crefrange{sec:method-margin-view-regression}{sec:method-margin-effective}, and margin learning directly models the relative distance between predicted score gaps $\Delta s_{ij}$ and mean opinion score (MOS) gaps $\Delta \mu_{ij}$.}
\label{fig:mr_reward}
\end{minipage}
\else
\begin{figure}[t]
\setlength{\abovecaptionskip}{2pt}
\setlength{\belowcaptionskip}{0pt}
\centering
\includegraphics[width=\columnwidth]{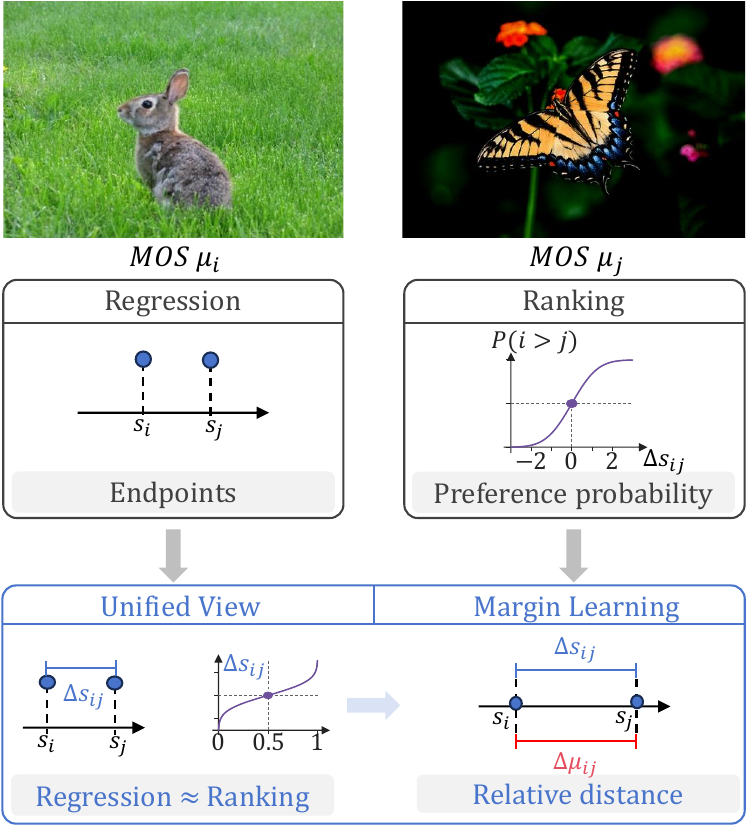}
\caption{Motivation of margin learning for BIQA. Regression estimates pointwise quality endpoints, while ranking maps pairwise differences into preference probabilities. Both methods can be interpreted through a unified margin view, as formalized in \crefrange{sec:method-margin-view-regression}{sec:method-margin-effective}, and margin learning directly models the relative distance between predicted score gaps $\Delta s_{ij}$ and mean opinion score (MOS) gaps $\Delta \mu_{ij}$.}
\label{fig:mr_reward}
\end{figure}
\fi

Despite this shift in representation framework, recent MLLM-based BIQA methods~\cite{q-align,deqa,q-insight,depictqa,q-instruct,Visualquality-r1,zoom-iqa} still rely on classical regression or ranking algorithms to define supervision. Regression provides a calibrated score target, but it also ties learning to dataset-specific score anchors. Pairwise ranking alleviates this issue by comparing images directly. In supervised fine-tuning (SFT)-based MLLM training, DeQA~\cite{deqa} combines score regression with a weighted Thurstone-style~\cite{thurstone1927law} fidelity loss to balance pointwise calibration and ordinal comparison. In reinforcement learning (RL)-based MLLM training, Q-Insight~\cite{q-insight} adopts a regression-style reward, while VisualQuality-R1 (VQ-R1)~\cite{Visualquality-r1} uses a Thurstone fidelity reward for pairwise ranking. Later work such as Zoom-IQA~\cite{zoom-iqa} also explores joint regression-ranking rewards. These designs are effective in practice, but the relation between regression and ranking remains largely empirical: their complementarity is usually adjusted by loss weights or reward design rather than explained by a shared optimization principle.

To explore this issue, we revisit regression and ranking under an RL training framework. For training images, the reward should admit a theoretical optimum. Regardless of whether supervision is derived from regression, ranking, or their combination, these objectives are eventually projected into the same reward space. This projection suggests that regression and ranking may share an underlying optimization target rather than acting as two independent signals.

As illustrated in \cref{fig:mr_reward}, both regression and ranking can be interpreted by the same relative quality distance, termed the \emph{quality margin}. At the objective level, regression fits margins induced by pointwise endpoints together with a dataset-anchor term, while Thurstone-style ranking~\cite{thurstone1927law} fits transformed margins through preference probabilities. We further connect this observation with the target metric of BIQA, showing that margin alignment bridges regression and ranking through the relational quality structure measured by PLCC. Based on this view, we propose \emph{MR-IQA}. Given a group of images, MR-IQA models the quality margin for every image pair within the group, suppressing the dataset-anchor interference inherited from regression, simplifying the ranking loss, and restoring the continuous distance information ignored by preference-only supervision.

Our main contributions are summarized as:
\begin{itemize}
\item We bridge regression and ranking with a unified quantity, quality margin, and theoretically derive why paradigms can be viewed as margin-oriented optimization.
\item We instantiate margin learning as \emph{MR-IQA}, a scale-controlled RL algorithm that directly evaluates whether predicted score gaps are underestimated, calibrated, or overestimated with respect to MOS margins.
\item We validate MR-IQA under controlled RL settings across six BIQA benchmarks, where it achieves the best average performance and provides a margin-based baseline for future RL-based BIQA studies.
\end{itemize}

Beyond empirical results, our derivation shows that quality margins connect score calibration with ordinal comparison through a unified relational structure. This view provides a foundation for BIQA and may also inform broader quality and ranking tasks such as aesthetic assessment, video quality assessment, and learning-to-rank.

\section{Related Work}
\label{sec:related}

\subsection{Regression-based BIQA}
\label{sec:related-regression}

CNNIQA~\cite{kang2014cnn} is an early representative showing that a CNN can learn no-reference quality prediction directly from image patches, while DeepBIQ~\cite{deepbiq} and DBCNN~\cite{dbcnn} further strengthened CNN-based score regression with deeper visual representations and bilinear feature modeling. NIMA~\cite{nima} takes a different route: it predicts a softmax distribution over ordered rating bins and derives the final quality score. MUSIQ~\cite{musiq} and MANIQA~\cite{yang2022maniqa} then extend score prediction with Transformer-based~\cite{dosovitskiy2020image} multi-scale and attention representations. Recent MLLM-based BIQA methods inherit this score-estimation view through label-wise supervision. Q-Align~\cite{q-align} uses discrete quality labels, and DeQA~\cite{deqa} extends one-hot labels to multi-label score distributions with cross-entropy supervision. With the shift to RL, Q-Insight~\cite{q-insight} converts label-wise score supervision into a continuous numerical reward for quality prediction. Regression-based method is easy to interpret, but remains sensitive to dataset-specific score anchors.

\subsection{Ranking-based BIQA}
\label{sec:related-ranking}

Gao et al.~\cite{gao2015learning} introduce learning-to-rank into BIQA, and dipIQ~\cite{dipiq} further learns from discriminable image pairs. RankIQA~\cite{rankiqa} constructs ranked examples from synthetically distorted images, while RRLW~\cite{gu2019rrlr} and CLRIQA~\cite{ou2020clriqa} extend the idea to recursive or controllable list-wise ranking. Transformer-based BIQA by Golestaneh et al.~\cite{golestaneh2022no} combines relative ranking with self-consistency, and later pairwise formulations such as rank-smoothed pairwise learning~\cite{rank_smoothed_pairwise} and PICNIQ~\cite{chahine2024picniq} further emphasize the usefulness of comparison labels. In MLLM-based BIQA, DeQA~\cite{deqa} introduces Thurstone-style~\cite{thurstone1927law} fidelity loss for ranking. VisualQuality-R1 (VQ-R1)~\cite{Visualquality-r1} extends this method to the RL region. But most ordinary ranking supervision mainly preserves preference direction and leaves the relative distance between images weakly specified.

\subsection{MLLM-based BIQA Training}
\label{sec:related-mllm}

MLLM-based BIQA~\cite{q-align, deqa, q-insight, depictqa, q-instruct, q-ponder,Visualquality-r1,h-iqa,zoom-iqa} is mainly trained through supervised fine-tuning (SFT) or reinforcement learning (RL). SFT methods such as Q-Align~\cite{q-align}, DeQA~\cite{deqa}, Q-Instruct~\cite{q-instruct}, and DepictQA~\cite{depictqa} adapt MLLMs with discrete labels, score-bin distributions, or curated instruction responses. This is effective for task adaptation, but when the training data lacks diversity, fixed targets can encourage template overfitting and weaken the original reasoning behavior of the model. RL instead computes rewards after sampling, allowing continuous values, pairwise relations, and reasoning behaviors to be supervised while retaining more of the base model's reasoning capability.

\section{Method}
\label{sec:method}

\begin{figure*}[t]
    \centering
    \includegraphics[width=\textwidth]{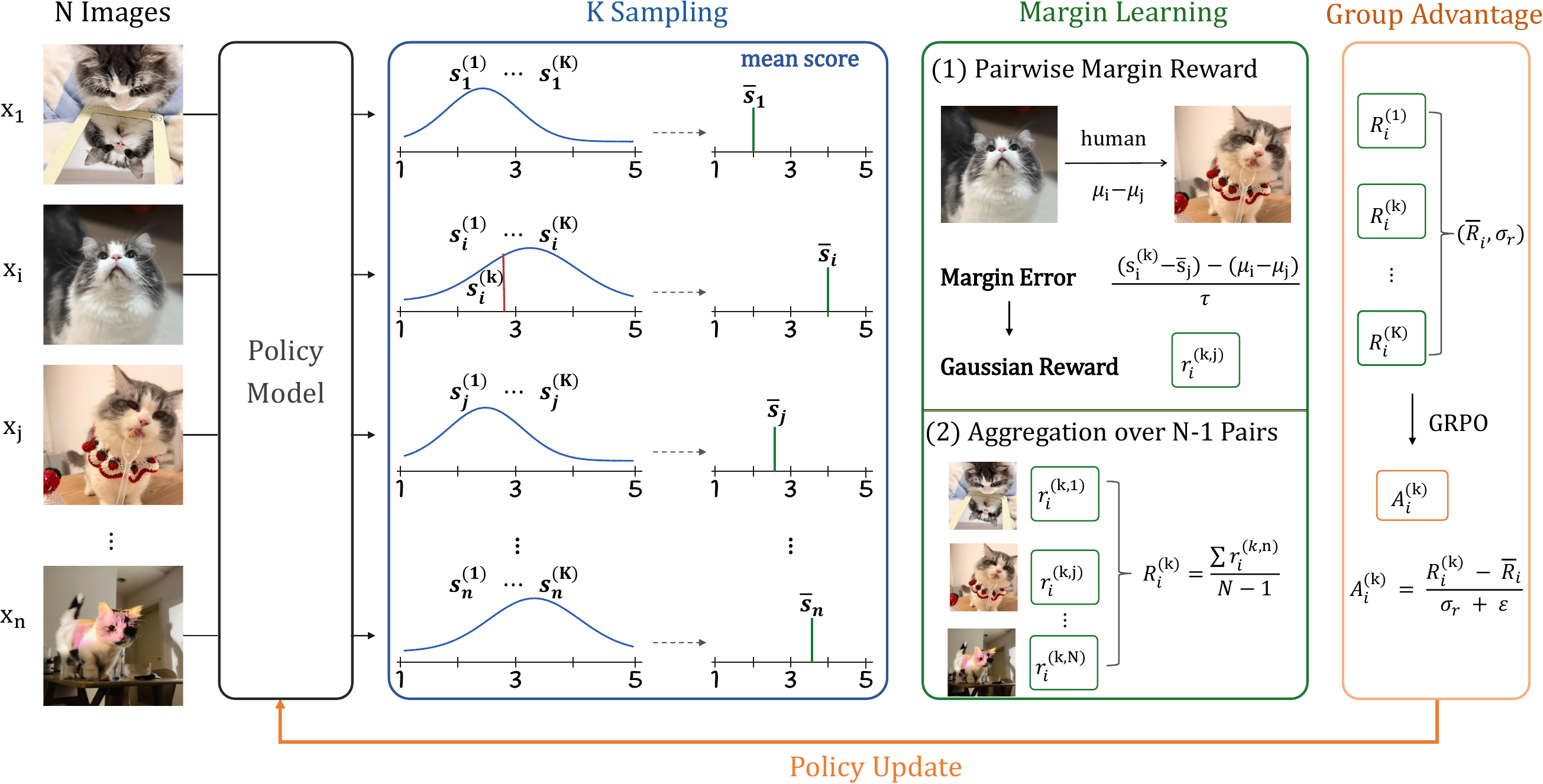}
    \caption{\textbf{MR-IQA training pipeline.} For a group of $N$ images, the policy model samples $K$ quality-score completions per image and forms image-level mean predictions. For one completion $s_i^{(k)}$, margin learning compares its predicted margin to the MOS margin against each other image, converts the margin error into a Gaussian pairwise reward, and aggregates the resulting $N{-}1$ rewards into $R_i^{(k)}$. Group Relative Policy Optimization (GRPO)~\cite{shao2024deepseekmath} then normalizes the sampled rewards into the advantage $A_i^{(k)}$ for policy update.}
    \label{fig:mr_pipeline}
\end{figure*}

In this section, we first revisit regression and ranking as two ways of optimizing margin-related objectives. We then explain why the margin acts as a bridge in relational quality structure. Finally, we instantiate margin learning as an RL algorithm that directly supervises pairwise margins.

\subsection{Definition of Quality Margins}
\label{sec:quality-margin-definition}

Following the statistical quality modeling view in DeQA~\cite{deqa} and VQ-R1~\cite{Visualquality-r1}, we regard the perceived quality of an image $x_i$ as a Gaussian variable and define the pairwise quality margin between two images $x_i$ and $x_j$ as
\begin{equation}
    x_i \sim \mathcal{N}(\mu_i,\sigma_i^2),
    \qquad
    \Delta\mu_{ij} = \mu_i-\mu_j.
    \label{eq:quality-margin-def}
\end{equation}
In the rest of this work, we use $\mu$ to denote human MOS, $\sigma$ to denote inter-rater standard deviation, and $s$ to denote model-sampled quality estimates.

\subsection{Margin View of Regression}
\label{sec:method-margin-view-regression}

In this part, we explain why pointwise regression has a strong connection with quality margins. For pointwise regression, each image has an endpoint error $e_i$; for quality margins, each image pair has a relational error $\delta_{ij}$, as defined in \crefrange{eq:discussion-centered-residual}{eq:discussion-margin-error}. We introduce the centered residual $\tilde{e}_i$ to make the dataset mean explicit:
\begin{align}
    \tilde{e}_i
    &=
    (s_i-\bar{s})-(\mu_i-\bar{\mu}),
    \label{eq:discussion-centered-residual}\\
    e_i
    &=
    s_i-\mu_i
    =
    \tilde{e}_i+(\bar{s}-\bar{\mu}),
    \label{eq:discussion-endpoint-error}\\
    \delta_{ij}
    &=
    \Delta s_{ij}-\Delta\mu_{ij}
    =
    \tilde{e}_i-\tilde{e}_j .
    \label{eq:discussion-margin-error}
\end{align}
Under an $L_2$ constraint, the accumulated errors of margin learning and pointwise regression can be decomposed as
\begin{align}
    \sum_{i<j}\delta_{ij}^{2}
    &=
    \sum_{i<j}(\tilde{e}_i-\tilde{e}_j)^2
    =
    N\sum_{i=1}^{N}\tilde{e}_i^2 ,
    \label{eq:discussion-margin-l2}\\
    \sum_{i=1}^{N} e_i^2
    &=
    \underbrace{
    \sum\nolimits_{i=1}^{N}\tilde{e}_i^2
    }_{
    \substack{\text{margin error}}
    }
    +
    \underbrace{N(\bar{s}-\bar{\mu})^2}_{\text{dataset-anchor}}
    .
    \label{eq:discussion-regression-l2}
\end{align}
Thus, pointwise regression can be viewed as optimizing two coupled parts: margin-error fitting and dataset-anchor learning. The global prediction shift $\bar{s}-\bar{\mu}$ can become restrictive under cross-dataset calibration shifts. Removing the dataset-anchor learning should benefit generalization.

\subsection{Margin View of Ranking}
\label{sec:method-margin-view-ranking}

In this part, we show the relation between ranking and margin learning by analyzing the optimization objective with a general ranking method as an example.
Thurstone-style ranking is a representative formulation in BIQA~\cite{thurstone1927law,deqa,Visualquality-r1}. Under Thurstone Case III~\cite{thurstone1927law}, the discriminal score associated with each image $x_i$ is modeled as an image-specific Gaussian distribution, and different images are assumed independent:
\begin{equation}
    z_{ij}
    =
    \frac{\Delta\mu_{ij}}{\sqrt{\sigma_i^2+\sigma_j^2}},
    \quad
    \Phi(z)
    =
    \int_{-\infty}^{z}
    \frac{e^{-\frac{t^2}{2}}}{\sqrt{2\pi}}\,dt .
    \label{eq:discussion-thurstone-z-cdf}
\end{equation}
\begin{equation}
    P_{ij}
    =
    \Phi\!\left(
    \frac{\Delta\mu_{ij}}{\sqrt{\sigma_i^2+\sigma_j^2}}
    \right),
    \label{eq:discussion-thurstone-probability}
\end{equation}
Here $z_{ij}$ is the normalized margin term, $\Phi(\cdot)$ is the standard normal cumulative distribution function, and $P_{ij}$ represents the label-side probability that $x_i$ has higher perceived quality than $x_j$ under the Thurstone model~\cite{thurstone1927law}.
Fidelity-based ranking models then compare the chosen label-side probability $P_{ij}$ with the prediction-side probability $\hat{P}_{ij}$. The fidelity-based ranking loss is
\begin{equation}
    \begin{aligned}
    \mathcal{L}_{\mathrm{fd},ij}
    &=
    1
    -
    \sqrt{P_{ij}\hat{P}_{ij}}
    -
    \sqrt{(1-P_{ij})(1-\hat{P}_{ij})} ,
    \end{aligned}
    \label{eq:discussion-fidelity-loss}
\end{equation}
\begin{equation}
    \begin{aligned}
    \frac{\partial \mathcal{L}_{\mathrm{fd},ij}}{\partial \hat{P}_{ij}}
    &=
    -\frac{1}{2}
    \sqrt{\frac{P_{ij}}{\hat{P}_{ij}}}
    +
    \frac{1}{2}
    \sqrt{\frac{1-P_{ij}}{1-\hat{P}_{ij}}}.
    \end{aligned}
    \label{eq:discussion-fidelity-gradient}
\end{equation}
From \cref{eq:discussion-fidelity-gradient}, the optimal solution of the fidelity loss is $\hat{P}_{ij}^{\star}=P_{ij}$. Since $\Phi(\cdot)$ in \cref{eq:discussion-thurstone-z-cdf} is strictly monotonic, it is invertible. The corresponding predicted and human margins can therefore be recovered as
\begin{equation}
    \begin{aligned}
    \Delta s_{ij}
    &=
    \sqrt{\hat{\sigma}_i^2+\hat{\sigma}_j^2}\,
    \Phi^{-1}\!\left(\hat{P}_{ij}\right),\\
    \Delta\mu_{ij}
    &=
    \sqrt{\sigma_i^2+\sigma_j^2}\,
    \Phi^{-1}\!\left(P_{ij}\right).
    \end{aligned}
    \label{eq:discussion-margin-inverse}
\end{equation}
Therefore, the optimization direction of fidelity ranking is similar to $\Delta s_{ij}\approx\Delta\mu_{ij}$.
Moreover, VQ-R1~\cite{Visualquality-r1} uses a hard label, which collapses the continuous probability into a discrete target and weaken label-side continuity. We discuss the variance-scale issue in \cref{app:interrater-variance,app:baseline-reward-definitions}. In this sense, Thurstone-style ranking also relies on normalized margins.

\providecommand{\bridgeRegressionRankingTitle}{Bridge Between Regression and Ranking}
\subsection{\bridgeRegressionRankingTitle}
\label{sec:method-margin-effective}

\Cref{sec:method-margin-view-regression,sec:method-margin-view-ranking} show that both regression and ranking are related to quality margins, but this relation alone is not sufficient to explain why margins form a bridge between them. We next show that, in the BIQA setting, the bridge role of margins comes from their direct connection to relational quality structure. Intuitively, both regression and ranking can serve BIQA as long as they recover a reliable relational quality structure. From a theoretical view, this structure is commonly evaluated by PLCC between estimated scores and human MOS. Given $N$ images, PLCC is defined as
\begin{equation}
    \mathrm{PLCC}(\mathbf{s},\boldsymbol{\mu})
    =
    \frac{
    \sum_{n=1}^{N}(s_n-\bar{s})(\mu_n-\bar{\mu})
    }{
    \sqrt{\sum_{n=1}^{N}(s_n-\bar{s})^2}
    \sqrt{\sum_{n=1}^{N}(\mu_n-\bar{\mu})^2}
    },
    \label{eq:plcc-original}
\end{equation}
where $\mathbf{s}=(s_1,\ldots,s_N)$, $\boldsymbol{\mu}=(\mu_1,\ldots,\mu_N)$, $\bar{s}$ and $\bar{\mu}$ denote their sample means, and $\Delta s_{ij}=s_i-s_j$ denotes the predicted margin for each image pair $i<j$.
The connection becomes exact after expanding the centered covariance term.\footnote{A detailed derivation of \cref{eq:plcc-cov-derivation} is provided in \cref{app:plcc-pairwise-proof}.} Specifically,
\begin{equation}
    \sum_{i<j}
    \Delta s_{ij}\Delta\mu_{ij}
    =
    N
    \sum_{n=1}^{N}
    (s_n-\bar{s})(\mu_n-\bar{\mu}).
    \label{eq:plcc-cov-derivation}
\end{equation}
The corresponding variance terms can be written as
\begin{align}
    \sum_{n=1}^{N}(s_n-\bar{s})^2
    &=
    \frac{1}{N}
    \sum_{i<j}
    (\Delta s_{ij})^2,
    \label{eq:plcc-var-s-pairwise}
    \\
    \sum_{n=1}^{N}(\mu_n-\bar{\mu})^2
    &=
    \frac{1}{N}
    \sum_{i<j}
    (\Delta\mu_{ij})^2.
    \label{eq:plcc-var-mu-pairwise}
\end{align}
Using \cref{eq:plcc-cov-derivation,eq:plcc-var-s-pairwise,eq:plcc-var-mu-pairwise} in \cref{eq:plcc-original} shows that PLCC is equivalent to the cosine similarity between predicted and human margin vectors:
\begin{equation}
    \begin{aligned}
    \mathrm{PLCC}(\mathbf{s},\boldsymbol{\mu})
    &=
    \frac{
    \sum_{i<j}\Delta s_{ij}\Delta\mu_{ij}
    }{
    \sqrt{\sum_{i<j}(\Delta s_{ij})^2}
    \sqrt{\sum_{i<j}(\Delta\mu_{ij})^2}
    }
    \\
    &=
    \operatorname{cosine}\!\left(
    \{\Delta s_{ij}\}_{i<j},
    \{\Delta\mu_{ij}\}_{i<j}
    \right).
    \end{aligned}
    \label{eq:plcc-pairwise}
\end{equation}

This equivalence theoretically explains the effectiveness of regression and ranking for BIQA. It also provides a new perspective: the underlying logic or optimization target of both regression and ranking is quality margin fitting. Therefore, directly modeling quality margins may provide a more direct and effective way to estimate quality structure.

\subsection{Modeling Pairwise Margin in BIQA}
\label{sec:method-mr}

We illustrate the concrete margin-learning pipeline in \cref{fig:mr_pipeline}. Consider a group of $N$ images $\mathcal{G}=\{x_i\}_{i=1}^{N}$, where each image $x_i$ is annotated with a MOS mean $\mu_i$. Following the margin definition in \cref{eq:quality-margin-def}, sufficient pairwise margins can describe the relative quality structure within the group.

For each image $x_i$, the policy samples $K$ completions and produces scalar quality scores $\{s_i^{(k)}\}_{k=1}^{K}$. We summarize the sampled scores of each image by their mean value:
\begin{equation}
    \bar{s}_i
    =
    \frac{1}{K}
    \sum_{k=1}^{K}s_i^{(k)}.
    \label{eq:model-mean}
\end{equation}

For the pairwise construction of completion $k$ from image $x_i$, we choose any comparison image $x_j$ with $j\ne i$ and compare the sampled score against the sampled mean of $x_j$:
\begin{equation}
    \begin{aligned}
    \Delta {s}_{ij}^{(k)} &= s_i^{(k)}-\bar{s}_j.
    \end{aligned}
    \label{eq:pairwise-margin-def}
\end{equation}
To separate margin modeling from the choice of error scale, we define a scale-controlled margin error:
\begin{equation}
    z_{ij}^{(k)}
    =
    \frac{
    \Delta {s}_{ij}^{(k)}-\Delta\mu_{ij}
    }{
    \tau_{ij}
    }.
    \label{eq:mr-normalized-error}
\end{equation}
where $\tau_{ij}$ is a positive scale term. We consider two choices: $\tau_{ij}^{\mathrm{raw}}=1$ and $\tau_{ij}^{\mathrm{unc}}=\sqrt{\sigma_i^2+\sigma_j^2}$.
The raw version penalizes the absolute mismatch between predicted and MOS margins, while the uncertainty-normalized version measures the mismatch relative to inter-rater disagreement.

We convert this error into reward with zero-centered margin estimators:
\begin{equation}
    r_{L_1,ij}^{(k)}=e^{-\left|z_{ij}^{(k)}\right|},\qquad
    r_{L_2,ij}^{(k)}=e^{-\frac{1}{2}\left(z_{ij}^{(k)}\right)^2}.
    \label{eq:margin-rewards}
\end{equation}
The $L_1$ form corresponds to a unit-scale Laplace likelihood with the constant factor omitted, while the $L_2$ estimator corresponds to a unit-variance Gaussian error model. The $L_1$ estimator is more robust to large normalized errors, while the $L_2$ estimator applies stronger pressure to large deviations; we use them as reward-design ablations.

\subsection{Image-Level Reward}
\label{sec:method-image-reward}

The pairwise identities in \cref{sec:method-margin-effective} use $i<j$ only to enumerate each unordered pair once. In the training pipeline, $x_i$ is the queried image, so completion $s_i^{(k)}$ is compared with every other image in the group; no ordering constraint is imposed, except that $j\ne i$.

For a group of $N$ images, each queried image $x_i$ forms $N{-}1$ pairwise comparisons with the other images. We aggregate these pairwise rewards directly into the final training reward:
\begin{equation}
    R_i^{(k)}
    =
    r_{\text{format},i}^{(k)}
    +
    \frac{1}{N-1}
    \sum_{\substack{j=1\\ j\ne i}}^{N}
    r_{\mathrm{margin},ij}^{(k)}.
    \label{eq:mr-reward}
\end{equation}
where $r_{\mathrm{margin},ij}^{(k)}$ can be instantiated by either the $L_1$ or $L_2$ estimator in \cref{eq:margin-rewards}. The format reward $r_{\text{format},i}^{(k)}$ checks whether the response follows the required answer format and whether the score can be parsed.

\subsection{Group Relative Policy Optimization}
\label{sec:method-grpo}

Following DeepSeek-Math~\cite{shao2024deepseekmath}, we use Group Relative Policy Optimization (GRPO) as the policy-update algorithm. GRPO converts the scalar rewards above into relative advantages for policy update. In one training rollout (\cref{fig:mr_pipeline}), the policy samples $K_i$ completions for each $x_i$, parses their scores, computes $R_i^{(k)}$ by comparing $s_i^{(k)}$ with $\bar{s}_j$ for all $j\ne i$, and then normalizes rewards across completions of the same image:
\begin{equation}
    A_i^{(k)}
    =
    \frac{
    R_i^{(k)}
    -
    \mathrm{mean}_{l}\!\left(R_i^{(l)}\right)}
    {\mathrm{std}_{l}\!\left(R_i^{(l)}\right)+\varepsilon},
    \label{eq:grpo-advantage}
\end{equation}
where $A_i^{(k)}$ is the advantage for completion $k$. A completion with above-average margin consistency receives a positive advantage and is reinforced, while a lower-reward completion is suppressed. With the importance ratio $\rho_i^{(k)}=\pi_\theta(o_i^{(k)}\mid x_i)/\pi_{\theta_{\text{old}}}(o_i^{(k)}\mid x_i)$ and its clipped version $\rho_{i,\mathrm{c}}^{(k)}=\mathrm{clip}(\rho_i^{(k)},1{-}\epsilon_{\mathrm{clip}},1{+}\epsilon_{\mathrm{clip}})$, the clipped GRPO objective is
\begin{equation}
\begin{aligned}
    \mathcal{L}_{\text{GRPO}}(\theta)
    =
    -\,\mathbb{E}\Big[
    &\min\!\left(
    \rho_i^{(k)} A_i^{(k)},
    \rho_{i,\mathrm{c}}^{(k)} A_i^{(k)}
    \right) \\
    &- \beta_{\text{KL}}\,
    \mathrm{KL}\!\left(\pi_\theta\,\|\,\pi_{\text{ref}}\right)
    \Big],
\end{aligned}
    \label{eq:grpo-loss}
\end{equation}
where $\epsilon_{\mathrm{clip}}$ is the GRPO clipping range and $\beta_{\text{KL}}$ controls reference-policy regularization. Larger values strengthen this regularization, trading adaptation for stability.

\section{Experiments}
\label{sec:exp}

\begin{table*}[t!]
\centering
\caption{\textbf{Main benchmark comparison.} All methods report PLCC$\uparrow$ and SRCC$\uparrow$ between predicted and ground-truth scores. \textcolor{bestred}{Red} numbers indicate overall best results; \textcolor{rlbestblue}{Blue} numbers indicate RL-block best results not already marked in red. Except for the reported Q-Insight~\cite{q-insight} row, RL-training models are reproduced under the controlled reproduction protocol with Qwen3-VL-2B~\cite{qwen3vl} as the backbone MLLM.}
\label{tab:main}
\footnotesize
\setlength{\tabcolsep}{2.5pt}
\resizebox{\textwidth}{!}{
\begin{tabular}{l cc cc cc cc cc cc cc}
\toprule
& \multicolumn{6}{c}{Authentic} & \multicolumn{2}{c}{AI-generated} & \multicolumn{4}{c}{Synthetic} & \multicolumn{2}{c}{Average} \\
\cmidrule(lr){2-7} \cmidrule(lr){8-9} \cmidrule(lr){10-13} \cmidrule(lr){14-15}
& \multicolumn{2}{c}{KonIQ} & \multicolumn{2}{c}{SPAQ} & \multicolumn{2}{c}{LIVE-W} & \multicolumn{2}{c}{AGIQA-3K} & \multicolumn{2}{c}{KADID-10k} & \multicolumn{2}{c}{CSIQ} & \multicolumn{2}{c}{Avg.} \\
\cmidrule(lr){2-3} \cmidrule(lr){4-5} \cmidrule(lr){6-7} \cmidrule(lr){8-9} \cmidrule(lr){10-11} \cmidrule(lr){12-13} \cmidrule(lr){14-15}
Method & PLCC & SRCC & PLCC & SRCC & PLCC & SRCC & PLCC & SRCC & PLCC & SRCC & PLCC & SRCC & PLCC & SRCC \\
\midrule
\multicolumn{15}{l}{\textit{Hand-crafted}} \\
NIQE~\cite{NIQE}      & 0.533 & 0.530 & 0.679 & 0.664 & 0.493 & 0.449 & 0.560 & 0.533 & 0.468 & 0.405 & 0.718 & 0.628 & 0.575 & 0.535 \\
BRISQUE~\cite{BRISQUE} & 0.225 & 0.226 & 0.490 & 0.406 & 0.361 & 0.313 & 0.541 & 0.497 & 0.429 & 0.356 & 0.740 & 0.556 & 0.464 & 0.392 \\
\midrule
\multicolumn{15}{l}{\textit{Deep-learning-based}} \\
NIMA~\cite{nima}                      & 0.896 & 0.859 & 0.838 & 0.856 & 0.814 & 0.771 & 0.715 & 0.654 & 0.532 & 0.535 & 0.695 & 0.649 & 0.748 & 0.721 \\
DBCNN~\cite{dbcnn}                    & 0.884 & 0.875 & 0.812 & 0.806 & 0.773 & 0.730 & 0.641 & 0.648 & 0.497 & 0.484 & 0.586 & 0.572 & 0.699 & 0.686 \\
MUSIQ~\cite{musiq}                    & 0.924 & 0.929 & 0.868 & 0.863 & 0.789 & 0.830 & 0.722 & 0.630 & 0.575 & 0.556 & 0.771 & 0.710 & 0.775 & 0.753 \\
MANIQA~\cite{yang2022maniqa}          & 0.849 & 0.834 & 0.768 & 0.758 & 0.849 & 0.832 & 0.723 & 0.636 & 0.499 & 0.465 & 0.623 & 0.627 & 0.719 & 0.692 \\
CLIP-IQA+~\cite{wang2023exploring}    & 0.909 & 0.895 & 0.866 & 0.864 & 0.832 & 0.805 & 0.736 & 0.685 & 0.653 & 0.654 & 0.772 & 0.719 & 0.795 & 0.770 \\
\midrule
\multicolumn{15}{l}{\textit{MLLM-based: SFT training}} \\
C2Score~\cite{zhu2024adaptive}            & 0.923 & 0.910 & 0.867 & 0.860 & 0.786 & 0.772 & 0.777 & 0.671 & 0.500 & 0.453 & 0.735 & 0.705 & 0.765 & 0.729 \\
Q-Align~\cite{q-align}                    & 0.941 & 0.940 & 0.886 & 0.887 & 0.853 & 0.860 & 0.772 & 0.735 & 0.674 & 0.684 & 0.671 & 0.737 & 0.800 & 0.807 \\
DeQA~\cite{deqa}                          & \textcolor{bestred}{0.953} & \textcolor{bestred}{0.941} & 0.895 & 0.896 & 0.892 & 0.879 & 0.809 & 0.729 & 0.694 & 0.687 & \textcolor{bestred}{0.787} & \textcolor{bestred}{0.744} & \textcolor{bestred}{0.838} & \textcolor{bestred}{0.813} \\
\multicolumn{15}{l}{\textit{MLLM-based: RL training}} \\
Q-Insight (regression)~\cite{q-insight}                & 0.918 & 0.895 & \textcolor{bestred}{0.903} & \textcolor{bestred}{0.903} & 0.870 & 0.839 & \textcolor{bestred}{0.816} & \textcolor{bestred}{0.766} & \textcolor{bestred}{0.702} & \textcolor{bestred}{0.702} & 0.685 & 0.640 & 0.816 & 0.791 \\

VQ-R1 (ranking)~\cite{Visualquality-r1}  & 0.886 & 0.919 & 0.867 & 0.887 & 0.817 & 0.869 & 0.744 & 0.718 & 0.635 & 0.640 & 0.709 & 0.721 & 0.776 & 0.792 \\

\rowcolor{oursrowgray}
\textbf{MR-IQA (ours)} & \textcolor{rlbestblue}{0.949} & \textcolor{rlbestblue}{0.931} & 0.892 & 0.897 & \textcolor{bestred}{0.899} & \textcolor{bestred}{0.883} & 0.804 & 0.732 & 0.672 & 0.683 & \textcolor{rlbestblue}{0.767} & \textcolor{rlbestblue}{0.732} & \textcolor{rlbestblue}{0.831} & \textcolor{rlbestblue}{0.810} \\

\bottomrule
\end{tabular}
}
\end{table*}

\subsection{Experimental Setup}
\label{sec:exp-setup}

\paragraph{Datasets.}
We train all controlled models only on the training split of KonIQ-10k~\cite{koniq}, which contains $7{,}046$ in-the-wild images at $512{\times}384$ resolution. Evaluation is performed on the KonIQ test split ($N{=}2{,}010$)~\cite{koniq} as the in-distribution authentic benchmark. For out-of-distribution (OOD) evaluation, we use authentic distortion datasets SPAQ ($N{=}11{,}125$)~\cite{spaq} and LIVE-Challenge ($N{=}1{,}169$)~\cite{live-w}, the AI-generated image quality dataset AGIQA-3K ($N{=}2{,}982$)~\cite{agiqa}, and synthetic distortion datasets KADID-10k ($N{=}10{,}125$)~\cite{kadid} and CSIQ ($N{=}866$)~\cite{csiq}.

\paragraph{Implementation details.}
We initialize the policy model from Qwen3-VL-2B~\cite{qwen3vl} and perform full-parameter fine-tuning with GRPO (\cref{eq:grpo-loss}). The backbone variants are drawn from Qwen3-VL~\cite{qwen3vl} and Qwen2.5-VL~\cite{qwen2.5}. We use AdamW~\cite{adam} with learning rate $1\times10^{-5}$, zero weight decay, and momentum parameters $\beta_1=0.9$, $\beta_2=0.999$. Runs use random seed $42$. For GRPO~\cite{shao2024deepseekmath}, we use $N=8$ images as a group, sample $K=6$ completions per image, set $\beta_{\mathrm{KL}}=0.02$, use temperature $0.7$, and perform four iterations per batch. Training runs for $10$ epochs on $8{\times}$ NVIDIA A6000 GPUs with per-device batch size $48$ generated completions; the per-epoch wall-clock time is approximately $57$ minutes for Qwen3-VL-2B~\cite{qwen3vl}, $1$ h $54$ min for Qwen3-VL-4B~\cite{qwen3vl}, and $2$ h for Qwen2.5-VL-7B~\cite{qwen2.5}.

\paragraph{Fair comparison protocol.}
Our controlled comparison focuses on general BIQA models under a fixed KonIQ-only training protocol~\cite{koniq}. Methods whose training pipelines rely on additional data, teacher distillation, non-matched protocols, or unavailable implementation code are treated as complementary literature rather than controlled baselines.

\begin{table*}[t!]
\centering
\caption{\textbf{Margin-reward ablation.} We use Qwen3-VL-2B~\cite{qwen3vl} as the ablation backbone and follow the same training and testing protocol as in \cref{sec:exp-setup}. The table evaluates combinations of $L_1$ or $L_2$ margin errors with two margin scales: $\tau_{ij}=1$ denotes the raw margin-error scale, and $\tau_{ij}^{\mathrm{unc}}=\sqrt{\sigma_i^2+\sigma_j^2}$ denotes the uncertainty-normalized scale. \textcolor{bestred}{Red} numbers indicate the best result for each metric. Overall, the $L_2$ variant with $\tau_{ij}=1$ achieves the best average PLCC/SRCC.}
\label{tab:ablation}
\footnotesize
\setlength{\tabcolsep}{2.1pt}
\renewcommand{\arraystretch}{0.88}
\resizebox{\textwidth}{!}{
\begin{tabular}{l cc cc cc cc cc cc cc}
\toprule
& \multicolumn{6}{c}{Authentic} & \multicolumn{2}{c}{AI-generated} & \multicolumn{4}{c}{Synthetic} & \multicolumn{2}{c}{Average} \\
\cmidrule(lr){2-7} \cmidrule(lr){8-9} \cmidrule(lr){10-13} \cmidrule(lr){14-15}
& \multicolumn{2}{c}{KonIQ} & \multicolumn{2}{c}{SPAQ} & \multicolumn{2}{c}{LIVE-W} & \multicolumn{2}{c}{AGIQA-3K} & \multicolumn{2}{c}{KADID-10k} & \multicolumn{2}{c}{CSIQ} & \multicolumn{2}{c}{Avg.} \\
\cmidrule(lr){2-3} \cmidrule(lr){4-5} \cmidrule(lr){6-7} \cmidrule(lr){8-9} \cmidrule(lr){10-11} \cmidrule(lr){12-13} \cmidrule(lr){14-15}
Variant & PLCC & SRCC & PLCC & SRCC & PLCC & SRCC & PLCC & SRCC & PLCC & SRCC & PLCC & SRCC & PLCC & SRCC \\
\midrule
Baseline & 0.685 & 0.669 & 0.780 & 0.797 & 0.717 & 0.722 & 0.733 & 0.691 & 0.565 & 0.551 & 0.696 & 0.707 & 0.696 & 0.690 \\
MR-IQA ($L_1$, $\tau_{ij}=1$) & 0.915 & 0.916 & 0.879 & 0.881 & 0.811 & 0.837 & 0.777 & 0.710 & 0.604 & 0.605 & 0.692 & 0.691 & 0.780 & 0.773 \\
MR-IQA ($L_1$, $\tau_{ij}^{\mathrm{unc}}$) & 0.934 & 0.922 & 0.875 & 0.891 & 0.870 & 0.851 & \textcolor{bestred}{0.808} & 0.715 & 0.557 & 0.601 & 0.736 & 0.705 & 0.797 & 0.781 \\
MR-IQA ($L_2$, $\tau_{ij}=1$) & \textcolor{bestred}{0.949} & \textcolor{bestred}{0.931} & \textcolor{bestred}{0.892} & \textcolor{bestred}{0.897} & \textcolor{bestred}{0.899} & \textcolor{bestred}{0.883} & 0.804 & \textcolor{bestred}{0.732} & \textcolor{bestred}{0.672} & \textcolor{bestred}{0.683} & \textcolor{bestred}{0.767} & \textcolor{bestred}{0.732} & \textcolor{bestred}{0.831} & \textcolor{bestred}{0.810} \\
MR-IQA ($L_2$, $\tau_{ij}^{\mathrm{unc}}$) & 0.933 & 0.923 & 0.883 & 0.885 & 0.853 & 0.849 & 0.800 & 0.711 & 0.602 & 0.631 & 0.730 & 0.712 & 0.800 & 0.785 \\
\bottomrule
\end{tabular}
}
\end{table*}

\begin{table*}[t!]
\centering
\caption{\textbf{Backbone stability.} Training and testing follow the protocol in \cref{sec:exp-setup}, except that Q-Insight with Qwen2.5-VL-7B follows its original settings~\cite{q-insight}. Backbone labels denote Qwen3-VL-2B/4B~\cite{qwen3vl} and Qwen2.5-VL-7B~\cite{qwen2.5}. For MR-IQA, we use the $L_2$ reward with $\tau_{ij}=1$. \protect\colorbox{oursrowgray}{Gray} rows mark our method, and \textcolor{bestred}{red} numbers indicate the best result under the same backbone.}
\label{tab:backbone-stability}
\scriptsize
\setlength{\tabcolsep}{2.0pt}
\renewcommand{\arraystretch}{0.88}
\resizebox{\textwidth}{!}{
\begin{tabular}{ll cc cc cc cc cc cc cc}
\toprule
& & \multicolumn{6}{c}{Authentic} & \multicolumn{2}{c}{AI-generated} & \multicolumn{4}{c}{Synthetic} & \multicolumn{2}{c}{Average} \\
\cmidrule(lr){3-8} \cmidrule(lr){9-10} \cmidrule(lr){11-14} \cmidrule(lr){15-16}
Backbone & Reward & \multicolumn{2}{c}{KonIQ} & \multicolumn{2}{c}{SPAQ} & \multicolumn{2}{c}{LIVE-W} & \multicolumn{2}{c}{AGIQA-3K} & \multicolumn{2}{c}{KADID-10k} & \multicolumn{2}{c}{CSIQ} & \multicolumn{2}{c}{Avg.} \\
\cmidrule(lr){3-4} \cmidrule(lr){5-6} \cmidrule(lr){7-8} \cmidrule(lr){9-10} \cmidrule(lr){11-12} \cmidrule(lr){13-14} \cmidrule(lr){15-16}
& & PLCC & SRCC & PLCC & SRCC & PLCC & SRCC & PLCC & SRCC & PLCC & SRCC & PLCC & SRCC & PLCC & SRCC \\
\midrule
Qwen3-2B & Baseline & 0.685 & 0.669 & 0.780 & 0.797 & 0.717 & 0.722 & 0.733 & 0.691 & 0.565 & 0.551 & 0.696 & 0.707 & 0.696 & 0.690 \\
             & Q-Insight & 0.934 & 0.923 & 0.883 & 0.886 & 0.872 & 0.863 & 0.800 & 0.731 & 0.652 & 0.673 & 0.759 & \textcolor{bestred}{0.752} & 0.817 & 0.805 \\
             & VQ-R1 & 0.886 & 0.919 & 0.867 & 0.887 & 0.817 & 0.869 & 0.744 & 0.718 & 0.635 & 0.640 & 0.709 & 0.721 & 0.776 & 0.792 \\
\rowcolor{oursrowgray}
             & \textbf{MR-IQA} & \textcolor{bestred}{0.949} & \textcolor{bestred}{0.931} & \textcolor{bestred}{0.892} & \textcolor{bestred}{0.897} & \textcolor{bestred}{0.899} & \textcolor{bestred}{0.883} & \textcolor{bestred}{0.804} & \textcolor{bestred}{0.732} & \textcolor{bestred}{0.672} & \textcolor{bestred}{0.683} & \textcolor{bestred}{0.767} & 0.732 & \textcolor{bestred}{0.831} & \textcolor{bestred}{0.810} \\
\midrule
Qwen3-4B & Baseline & 0.748 & 0.688 & 0.873 & 0.875 & 0.771 & 0.757 & 0.794 & \textcolor{bestred}{0.737} & 0.649 & 0.666 & \textcolor{bestred}{0.785} & \textcolor{bestred}{0.750} & 0.770 & 0.745 \\
             & Q-Insight & 0.934 & 0.918 & 0.886 & 0.888 & 0.868 & 0.857 & \textcolor{bestred}{0.806} & 0.720 & 0.590 & 0.618 & 0.724 & 0.701 & 0.801 & 0.784 \\
             & VQ-R1 & 0.931 & 0.924 & 0.886 & 0.889 & 0.849 & \textcolor{bestred}{0.873} & 0.779 & 0.710 & 0.627 & 0.642 & 0.673 & 0.683 & 0.791 & 0.787 \\
\rowcolor{oursrowgray}
             & \textbf{MR-IQA} & \textcolor{bestred}{0.945} & \textcolor{bestred}{0.927} & \textcolor{bestred}{0.893} & \textcolor{bestred}{0.895} & \textcolor{bestred}{0.876} & 0.870 & 0.793 & 0.710 & \textcolor{bestred}{0.670} & \textcolor{bestred}{0.695} & 0.758 & 0.747 & \textcolor{bestred}{0.823} & \textcolor{bestred}{0.807} \\
\midrule
Qwen2.5-7B & Baseline & 0.741 & 0.684 & 0.872 & 0.871 & 0.772 & 0.742 & 0.811 & 0.753 & 0.601 & 0.581 & 0.699 & 0.652 & 0.750 & 0.714 \\
               & Q-Insight & 0.920 & 0.907 & \textcolor{bestred}{0.886} & 0.884 & 0.878 & 0.851 & 0.806 & \textcolor{bestred}{0.767} & 0.629 & \textcolor{bestred}{0.699} & 0.687 & 0.634 & 0.801 & 0.790 \\
               & VQ-R1 & 0.897 & 0.921 & 0.849 & \textcolor{bestred}{0.891} & 0.761 & 0.858 & 0.774 & 0.754 & 0.662 & 0.675 & 0.517 & 0.605 & 0.743 & 0.784 \\
\rowcolor{oursrowgray}
               & \textbf{MR-IQA} & \textcolor{bestred}{0.950} & \textcolor{bestred}{0.936} & 0.861 & 0.862 & \textcolor{bestred}{0.888} & \textcolor{bestred}{0.877} & \textcolor{bestred}{0.818} & 0.756 & \textcolor{bestred}{0.669} & 0.669 & \textcolor{bestred}{0.702} & \textcolor{bestred}{0.681} & \textcolor{bestred}{0.815} & \textcolor{bestred}{0.797} \\
\bottomrule
\end{tabular}
}
\end{table*}

\subsection{Main Results}
\label{sec:exp-main}

\textbf{Compared model families.} \cref{tab:main} compares MR-IQA with representative BIQA families. The hand-crafted group includes NIQE~\cite{NIQE} and BRISQUE~\cite{BRISQUE}; the deep-learning-based group includes NIMA~\cite{nima}, DBCNN~\cite{dbcnn}, MUSIQ~\cite{musiq}, MANIQA~\cite{yang2022maniqa}, and CLIP-IQA+~\cite{wang2023exploring}. For MLLM-based BIQA, we separate SFT methods, including C2Score~\cite{zhu2024adaptive}, Q-Align~\cite{q-align}, and DeQA~\cite{deqa}, from RL methods, including Q-Insight~\cite{q-insight}, and VQ-R1~\cite{Visualquality-r1}. \\
\textbf{Competitive overall performance.} Across all methods, MR-IQA reaches an average PLCC/SRCC of $0.831/0.810$, which is close to the strong SFT-based model DeQA~\cite{deqa} at $0.838/0.813$. This gap is notable because MR-IQA remains an RL-based model and can preserve response-level quality reasoning behavior. We argue that quality margin learning has the potential to approach DeQA~\cite{deqa} with delicate optimization, because DeQA~\cite{deqa} adopts a joint loss.\\
\textbf{RL-based comparison.} Within the RL-training block, Q-Insight~\cite{q-insight} is included as the regression representative using its reported Qwen2.5-VL-7B~\cite{qwen2.5} performance, while VQ-R1~\cite{Visualquality-r1} is included as the ranking representative. VQ-R1 and MR-IQA are evaluated under the controlled Qwen3-VL-2B~\cite{qwen3vl} reproduction because the original VQ-R1 protocol is not matched. MR-IQA obtains the best average RL performance. At the same time, Q-Insight remains stronger on AGIQA-3K~\cite{agiqa} and KADID-10k~\cite{kadid}. Since these gaps may be affected by backbone and training environment differences, \cref{sec:exp-backbone} further compares regression, ranking, and margin learning under matched backbone settings.

\subsection{Ablation Study}
\label{sec:exp-ablation}

\textbf{Reward function.}
Under the Qwen3-VL-2B~\cite{qwen3vl} baseline, we compare the $L_1$ and $L_2$ margin estimators with two scale choices: the raw margin scale $\tau_{ij}=1$ and the human-uncertainty scale $\tau_{ij}^{\mathrm{unc}}=\sqrt{\sigma_i^2+\sigma_j^2}$. The raw-margin $L_2$ design achieves the best overall performance, with average PLCC/ SRCC gains of $0.135/0.120$ over the baseline. In contrast, human-uncertainty normalization does not consistently improve the results, suggesting that inter-rater variance is not always a reliable training-time margin scale.\\

\subsection{Backbone and Algorithm Stability}
\label{sec:exp-backbone}

To examine whether the gap between MR-IQA and regression/ranking algorithms depends on a specific backbone, \cref{tab:backbone-stability} evaluates the algorithm families on Qwen3-VL-2B/4B~\cite{qwen3vl} and Qwen2.5-VL-7B~\cite{qwen2.5}. Qwen2.5-VL-7B~\cite{qwen2.5} is the backbone used by the original Q-Insight~\cite{q-insight} and VQ-R1~\cite{Visualquality-r1} settings. Except for the Qwen2.5-VL-7B~\cite{qwen2.5} Q-Insight~\cite{q-insight} run, which follows its official training script, the controlled RL rows use matched data and hyperparameter settings under the protocol in \cref{sec:exp-setup}. Under these controlled settings, margin learning achieves the strongest average PLCC/SRCC across the tested backbones, indicating that its gains over regression and ranking algorithms are not tied to a single model scale nor a specific backbone.\footnote{Additional ablation studies and training dynamics are provided in Appendix~\cref{app:additional-ablation,app:convergence-efficiency}.}

\begin{figure*}[t]
    \centering
    \includegraphics[width=\textwidth]{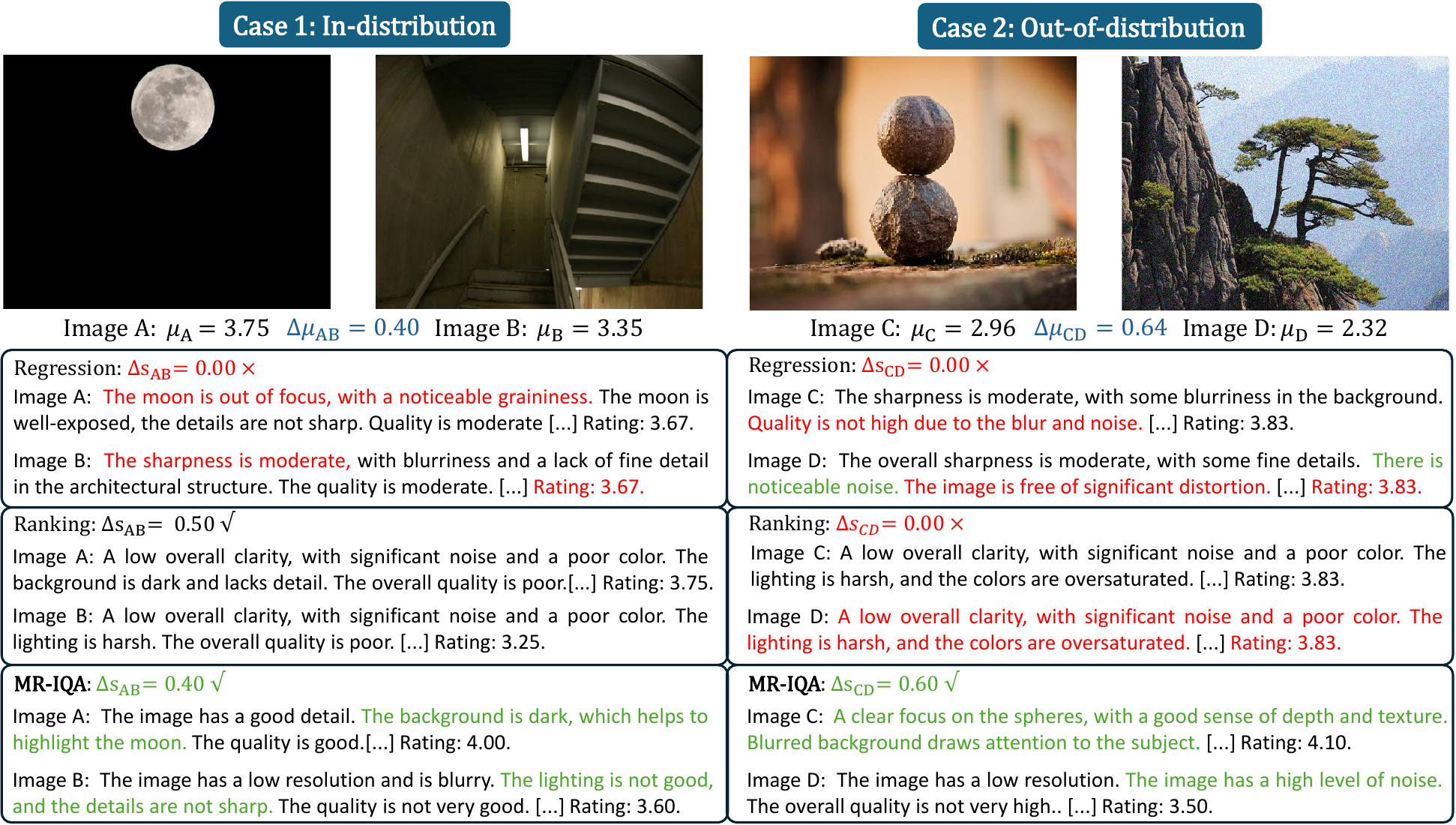}
    \caption{\textbf{Qualitative case study of three algorithms.} We compare reproduced Q-Insight regression~\cite{q-insight}, VQ-R1 ranking~\cite{Visualquality-r1}, and MR-IQA (ours), all using Qwen3-VL-2B~\cite{qwen3vl} as the backbone. The in-distribution examples are sampled from KonIQ~\cite{koniq}, and the out-of-distribution examples are sampled from KADID-10k~\cite{kadid}. \textcolor{red}{Red} highlights potential perceptual or scoring errors, while \textcolor{rlbestgreen}{green} marks correct perceptual evidence. Overall, MR-IQA aligns visual evidence and MOS margins more robustly across distributions, whereas the ranking baseline shows signs of perceptual overfitting on the out-of-distribution samples.}
    \label{fig:case-study-v3}
\end{figure*}

\subsection{Case Study}
\label{sec:exp-case-study}

\Cref{fig:case-study-v3} compares Q-Insight (regression)~\cite{q-insight}, VQ-R1 (ranking)~\cite{Visualquality-r1}, and MR-IQA on in- and out-of-distribution samples. MR-IQA may show dataset-anchor shifts without pointwise MOS regression, but it estimates relative margins more accurately than both baselines and identifies correct perceptual evidence earlier.

\section{Discussion}
\label{sec:discussion}

\subsection*{Why variance not always helpful?}
\label{sec:discussion-interrater-variance}

In the Thurstone model~\cite{thurstone1927law}, variance characterizes the instability of an image's perceived quality distribution. In BIQA, this instability is usually observed as inter-rater variance. A natural expectation is therefore that variance should help calibrate margin learning by down-weighting uncertain image pairs. However, our controlled ablation in \cref{tab:ablation} shows that using inter-rater variance as the normalization scale does not consistently improve performance. In other Thurstone-style models, DeQA~\cite{deqa} uses a fixed variance during training, whereas VQ-R1~\cite{Visualquality-r1} uses model sampling variance rather than human variance. \\
\textbf{Possible reasons.} \textbf{(1)} There is a human-model mismatch. Human inter-rater variance can be large because individuals have different perceptual preferences and annotation habits. Model sampling variance is usually much smaller because stable score generation is itself a desirable behavior for BIQA models. \textbf{(2)} Model variance may not be fully captured by repeated sampled scalar outputs. A more faithful uncertainty estimate may need token-level alternatives. \textbf{(3)} Dataset-level variance behavior differs strongly; see Appendix~\cref{app:interrater-variance}. \textbf{(4)} The standard BIQA metrics, PLCC and SRCC, evaluate mean score prediction and rank order rather than predictive variance. From the evaluation perspective, variance may describe reliability or ambiguity, but it is not directly rewarded by the metrics.

\section{Limitations and Future Work}
\label{sec:limit}

This work has several limitations. First, inter-rater variance did not consistently improve margin learning, although uncertainty may still encode reliability or ambiguity. Second, experiments mainly use Qwen-family MLLMs; testing other MLLMs and non-MLLM backbones is needed for generality and efficiency. Third, margin learning needs sufficient pair coverage and sampled completions, so small datasets may yield noisy margins. Finally, we do not deeply analyze visual-quality reasoning; future work should examine reasoning traces, and attribute-level explanations.

\raggedbottom
\section{Conclusion}
\label{sec:conclusion}

This work revisits regression and ranking in BIQA through the lens of quality margins. We show that pointwise regression fits pairwise margins together with a dataset-anchor term, while Thurstone-style ranking fits transformed margins through preference probabilities. Together with the pairwise form of PLCC, this reveals margin fitting as a shared mechanism behind score calibration and ordinal comparison. Guided by this view, we propose MR-IQA, an RL-based framework that directly rewards calibrated MOS margins from sampled scores. Across general BIQA benchmarks and controlled comparisons, MR-IQA achieves competitive overall performance. Overall, this unified margin view explains why previous joint regression and ranking designs can be effective and provides a foundation for future optimization methods that directly model quality structure. It further turns the empirical complementarity between score calibration and ordinal comparison into an explicit optimization principle for future BIQA design.

\clearpage
\flushbottom

{
    \small
    \bibliographystyle{ieeenat_fullname}
    \bibliography{main}
}

\clearpage
\appendix
\setcounter{equation}{0}
\setcounter{figure}{0}
\setcounter{table}{0}
\renewcommand{\theHequation}{appendix.\arabic{equation}}
\renewcommand{\theHfigure}{appendix.\arabic{figure}}
\renewcommand{\theHtable}{appendix.\arabic{table}}
\twocolumn[{
\begin{center}
{\Large\bfseries Supplementary Material\par}
\vspace{0.3em}
{\Large\bfseries MR-IQA: A Unified Margin View of Regression and Ranking for Blind Image Quality Assessment\par}
\end{center}
\vspace{3.0em}
}]
\raggedbottom
\setlength{\textfloatsep}{8pt plus 2pt minus 2pt}
\setlength{\floatsep}{6pt plus 2pt minus 2pt}
\setlength{\intextsep}{6pt plus 2pt minus 2pt}
\setlength{\abovecaptionskip}{3pt}
\setlength{\belowcaptionskip}{0pt}

\section{Proof of the Margin Identity}
\label{app:plcc-pairwise-proof}

We provide the derivation of the pairwise margin-covariance identity in Eq.~(13) of the main paper. Let
\begin{equation}
    T
    =
    \sum_{i=1}^{N}\sum_{j=1}^{N}
    (s_i-s_j)(\mu_i-\mu_j).
    \label{eq:app-pairwise-t}
\end{equation}
Each unordered pair $(i,j)$ with $i<j$ appears twice in the ordered summation, as $(i,j)$ and $(j,i)$, and the two terms have the same product. Therefore,
\begin{equation}
    T
    =
    2\sum_{i<j}
    (s_i-s_j)(\mu_i-\mu_j)
    =
    2\sum_{i<j}
    \Delta s_{ij}\Delta\mu_{ij}.
    \label{eq:app-unordered-to-ordered}
\end{equation}
On the other hand, expanding $T$ gives
\begin{align}
    T
    &=
    \sum_{i=1}^{N}\sum_{j=1}^{N}
    \left(
    s_i\mu_i
    -
    s_i\mu_j
    -
    s_j\mu_i
    +
    s_j\mu_j
    \right)
    \notag\\
    &=
    N\sum_{i=1}^{N}s_i\mu_i
    -
    \left(\sum_{i=1}^{N}s_i\right)
    \left(\sum_{j=1}^{N}\mu_j\right)
    \notag\\
    &\quad
    -
    \left(\sum_{j=1}^{N}s_j\right)
    \left(\sum_{i=1}^{N}\mu_i\right)
    +
    N\sum_{j=1}^{N}s_j\mu_j
    \notag\\
    &=
    2N\sum_{n=1}^{N}s_n\mu_n
    -
    2N^2\bar{s}\bar{\mu}.
    \label{eq:app-t-expanded}
\end{align}
The centered covariance term satisfies
\begin{align}
    &\sum_{n=1}^{N}
    (s_n-\bar{s})
    (\mu_n-\bar{\mu})
    \notag\\
    &=
    \sum_{n=1}^{N}s_n\mu_n
    -
    \bar{\mu}\sum_{n=1}^{N}s_n
    -
    \bar{s}\sum_{n=1}^{N}\mu_n
    +
    N\bar{s}\bar{\mu}
    \notag\\
    &=
    \sum_{n=1}^{N}s_n\mu_n
    -
    N\bar{s}\bar{\mu}.
    \label{eq:app-centered-cov}
\end{align}
Combining \cref{eq:app-t-expanded,eq:app-centered-cov} yields
\begin{equation}
    T
    =
    2N
    \sum_{n=1}^{N}
    (s_n-\bar{s})(\mu_n-\bar{\mu}).
    \label{eq:app-t-centered}
\end{equation}
Finally, combining \cref{eq:app-t-centered} with \cref{eq:app-unordered-to-ordered} gives
\begin{equation}
    \sum_{i<j}
    (s_i-s_j)(\mu_i-\mu_j)
    =
    N
    \sum_{n=1}^{N}
    (s_n-\bar{s})(\mu_n-\bar{\mu}).
    \label{eq:app-pairwise-cov-identity}
\end{equation}
\section{Ablation of Group and Sampling Size}
\label{app:additional-ablation}

\label{app:nk-ablation}
We further investigate the effect of group size $N$ and sampling number $K$.
The $N$/$K$ rows of \cref{tab:app-nk-ablation} use the same Qwen3-VL-2B~\cite{qwen3vl} backbone, and the same training protocol as in main-paper.

\begin{table*}[t!]
\centering
\caption{\textbf{Controlled ablations and epoch generalization.} The $N$/$K$ rows vary only the group size $N$ or sampling number $K$ under the protocol of main-paper Table~2; \textcolor{bestred}{red} marks the best performance within each ablation block among available results. The checkpoint rows compare the epoch-3 and epoch-10 checkpoints for the default $N=8,K=6$ setting; $\Delta$ reports epoch 10 minus epoch 3.}
\label{tab:app-nk-ablation}
\scriptsize
\setlength{\tabcolsep}{1.7pt}
\renewcommand{\arraystretch}{0.88}
\resizebox{\textwidth}{!}{
\begin{tabular}{cc cc cc cc cc cc cc cc}
\toprule
& & \multicolumn{6}{c}{Authentic} & \multicolumn{2}{c}{AI-generated} & \multicolumn{4}{c}{Synthetic} & \multicolumn{2}{c}{Average} \\
\cmidrule(lr){3-8} \cmidrule(lr){9-10} \cmidrule(lr){11-14} \cmidrule(lr){15-16}
$N$ & $K$ & \multicolumn{2}{c}{KonIQ} & \multicolumn{2}{c}{SPAQ} & \multicolumn{2}{c}{LIVE-W} & \multicolumn{2}{c}{AGIQA-3K} & \multicolumn{2}{c}{KADID-10k} & \multicolumn{2}{c}{CSIQ} & \multicolumn{2}{c}{Avg.} \\
\cmidrule(lr){3-4} \cmidrule(lr){5-6} \cmidrule(lr){7-8} \cmidrule(lr){9-10} \cmidrule(lr){11-12} \cmidrule(lr){13-14} \cmidrule(lr){15-16}
& & PLCC & SRCC & PLCC & SRCC & PLCC & SRCC & PLCC & SRCC & PLCC & SRCC & PLCC & SRCC & PLCC & SRCC \\
\midrule
8 & 6 & \textcolor{bestred}{0.949} & \textcolor{bestred}{0.931} & 0.892 & \textcolor{bestred}{0.897} & \textcolor{bestred}{0.899} & \textcolor{bestred}{0.883} & 0.804 & \textcolor{bestred}{0.732} & \textcolor{bestred}{0.672} & \textcolor{bestred}{0.683} & 0.767 & 0.732 & \textcolor{bestred}{0.831} & \textcolor{bestred}{0.810} \\
6 & 6 & 0.945 & 0.929 & \textcolor{bestred}{0.893} & 0.896 & 0.887 & 0.877 & 0.805 & 0.713 & 0.648 & 0.655 & 0.768 & \textcolor{bestred}{0.749} & 0.824 & 0.803 \\
4 & 6 & 0.944 & 0.927 & 0.890 & 0.895 & 0.880 & 0.868 & \textcolor{bestred}{0.806} & 0.726 & 0.646 & 0.658 & \textcolor{bestred}{0.775} & 0.742 & 0.823 & 0.803 \\
\midrule
6 & 8 & 0.911 & 0.901 & 0.877 & 0.876 & 0.801 & 0.823 & 0.779 & 0.706 & 0.604 & 0.603 & 0.714 & 0.723 & 0.781 & 0.772 \\
6 & 6 & \textcolor{bestred}{0.945} & \textcolor{bestred}{0.929} & \textcolor{bestred}{0.893} & \textcolor{bestred}{0.896} & \textcolor{bestred}{0.887} & \textcolor{bestred}{0.877} & \textcolor{bestred}{0.805} & \textcolor{bestred}{0.713} & \textcolor{bestred}{0.648} & \textcolor{bestred}{0.655} & \textcolor{bestred}{0.768} & \textcolor{bestred}{0.749} & \textcolor{bestred}{0.824} & \textcolor{bestred}{0.803} \\
6 & 4 & 0.938 & 0.923 & 0.890 & 0.891 & 0.871 & 0.860 & 0.799 & 0.708 & 0.608 & 0.626 & 0.768 & 0.749 & 0.812 & 0.793 \\
\midrule[0.8pt]
\multicolumn{2}{l}{Epoch 3} & 0.944 & 0.931 & 0.892 & 0.897 & 0.890 & 0.878 & 0.804 & 0.722 & 0.620 & 0.656 & 0.732 & 0.697 & 0.814 & 0.797 \\
\multicolumn{2}{l}{Epoch 10} & 0.949 & 0.931 & 0.892 & 0.897 & 0.899 & 0.883 & 0.804 & 0.732 & 0.672 & 0.683 & 0.767 & 0.732 & 0.831 & 0.810 \\
\midrule
\multicolumn{2}{l}{$\Delta$} & \textcolor{rlbestgreen}{+0.005} & +0.000 & +0.000 & +0.000 & \textcolor{rlbestgreen}{+0.009} & \textcolor{rlbestgreen}{+0.005} & +0.000 & \textcolor{rlbestgreen}{+0.010} & \textcolor{rlbestgreen}{+0.052} & \textcolor{rlbestgreen}{+0.027} & \textcolor{rlbestgreen}{+0.035} & \textcolor{rlbestgreen}{+0.035} & \textcolor{rlbestgreen}{+0.017} & \textcolor{rlbestgreen}{+0.013} \\
\bottomrule
\end{tabular}
}
\end{table*}

Overall, the setting with sampling number $K=6$ and a group size of $N=8$ images achieves the best average performance.
When $K$ is fixed at $6$, a larger group size appears more favorable within the tested range.
When $N$ is fixed at $6$, $K=6$ performs best, but the effect of sampling number does not show a clear monotonic trend.
Sweeping $K$ under a larger group size such as $N=8$ is not computationally affordable for us and is left for future study.

\section{Training Dynamics}
\label{app:convergence-efficiency}

\begin{figure*}[t]
    \centering
    \includegraphics[width=\textwidth]{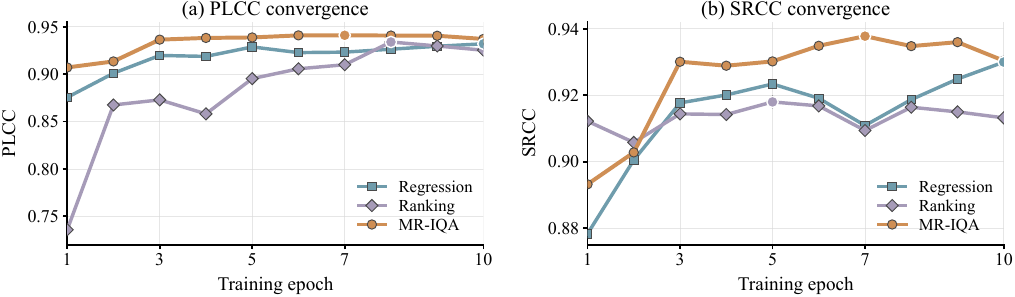}
    \caption{Convergence curves on a randomly sampled $200$-image in-domain KonIQ~\cite{koniq} diagnostic subset. Curves report PLCC$\uparrow$ and SRCC$\uparrow$ after each training epoch.}
    \label{fig:training-dynamics}
\end{figure*}

We examine convergence behavior on a diagnostic subset of $200$ randomly sampled in-domain KonIQ~\cite{koniq} images to visualize optimization dynamics. \Cref{fig:training-dynamics} shows that MR-IQA converges faster in the in-domain setting. Compared with the ranking baseline, this behavior is reasonable because margin learning preserves scale information in addition to ordinal direction. More interestingly, MR-IQA also rises faster than the regression baseline. One possible explanation is that regression observes each image as an isolated target, whereas a margin reward compares each image with multiple peers in the same group and therefore exposes richer relational supervision per update.

The checkpoint rows in \cref{tab:app-nk-ablation} further compare the epoch-3 checkpoint with the epoch-10 checkpoint under the same cross-benchmark protocol. The epoch-3 checkpoint already performs close to the final checkpoint on several benchmarks, suggesting that the model reaches a strong solution early. Continued training still brings modest average gains of $+0.017$ PLCC and $+0.013$ SRCC, with larger improvements on synthetic datasets. This suggests that later epochs primarily refine cross-dataset generalization rather than changing the learned in-domain quality structure.

\section{Inter-rater Variance Analysis}
\label{app:interrater-variance}

\begin{figure*}[t]
    \centering
    \includegraphics[width=0.93\textwidth]{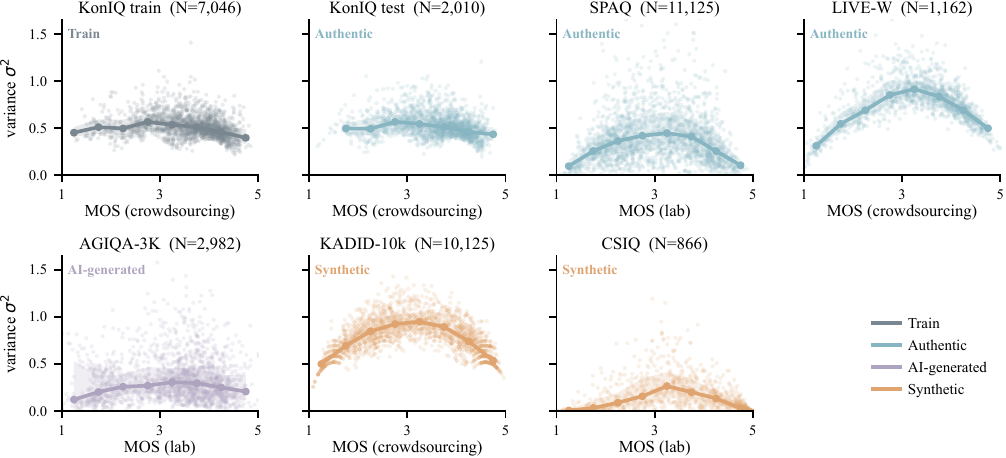}
    \caption{MOS-conditioned inter-rater variance distributions for KonIQ~\cite{koniq}, SPAQ~\cite{spaq}, LIVE-W~\cite{live-w}, AGIQA-3K~\cite{agiqa}, KADID-10k~\cite{kadid}, and CSIQ~\cite{csiq}. Axis suffixes denote the collection protocol: crowdsourcing or lab for controlled laboratory studies. Points denote samples, curves denote binned medians, and bands denote interquartile ranges. The dataset-dependent scale and shape suggest that manifest-level variance is informative but imperfect as a proxy for human uncertainty.}
    \label{fig:app-interrater-variance}
\end{figure*}

We further audit the per-image human disagreement statistics available in the current data manifests. To ensure consistency with the training and evaluation settings, and to keep the score scales comparable across datasets, we use the normalized statistics provided by DeQA~\cite{deqa}. \Cref{fig:app-interrater-variance} visualizes the MOS-conditioned variance distributions for KonIQ~\cite{koniq}, SPAQ~\cite{spaq}, LIVE-W~\cite{live-w}, AGIQA-3K~\cite{agiqa}, KADID-10k~\cite{kadid}, and CSIQ~\cite{csiq}.

The statistics help explain why directly using manifest variance as a human-uncertainty proxy can be fragile. First, the absolute variance scale is strongly dataset-dependent: LIVE-W~\cite{live-w} and KADID-10k~\cite{kadid} have much larger variance than KonIQ~\cite{koniq}, whereas CSIQ~\cite{csiq} has substantially lower variance. When such values are placed in the denominator of a normalized margin error, the reward becomes less sensitive on high-variance datasets and overly sharp on low-variance datasets, even if the downstream evaluation only uses MOS means. Second, the MOS-conditioned variance pattern is not stable: some datasets show higher variance around middle quality levels, while others exhibit flatter or lower-variance distributions. Thus, a large annotated variance does not always mean that a pairwise quality relation should be down-weighted in the same way across datasets.

These observations do not imply that rater variance is useless. Rather, they suggest that the current field mixes multiple factors, including observer disagreement, dataset protocol, score normalization, and possibly content difficulty. This can explain why variance-aware normalization may underperform or behave inconsistently compared with the variant without variance in main-paper Table~2. A more reliable use of human uncertainty may require dataset-specific calibration, robust clipping, or richer annotation models beyond a Gaussian standard deviation.

\section{Additional Reward Definitions}
\label{app:baseline-rewards}

\subsection{Baseline Reward Definitions}
\label{app:baseline-reward-definitions}

For the Q-Insight baseline~\cite{q-insight}, we follow the official fixed-threshold binary reward:
\begin{equation}
    r_{\mathrm{QI},i}^{(k)}
    =
    \mathbf{1}\!\left[
    |s_i^{(k)}-\mu_i|\le \tau
    \right],
    \qquad
    \tau=0.35.
    \label{eq:qinsight-reward}
\end{equation}
The continuous Thurstone preference probability~\cite{thurstone1927law} is
\begin{equation}
    p(i>j)
    =
    \Phi\!\left(
    \frac{\mu_i-\mu_j}
    {\sqrt{\sigma_i^2+\sigma_j^2}}
    \right).
    \label{eq:thurstone}
\end{equation}

For the VQ-R1 baseline~\cite{Visualquality-r1}, we follow the official Thurstone-style~\cite{thurstone1927law} fidelity reward. Here ``Thurstone-style'' refers to the preference-modeling principle, while the implemented reward is prediction-dependent and uses the model prediction sampling variance. Let $\hat{\sigma}_i^2$ and $\hat{\sigma}_j^2$ denote the model prediction sampling variances. The predicted preference probability for completion $k$ of image $x_i$ is
\begin{equation}
    p_{ij}^{(k)}
    =
    \Phi\!\left(
    \frac{s_i^{(k)}-\bar{s}_j}
    {\sqrt{\hat{\sigma}_i^2+\hat{\sigma}_j^2}}
    \right).
    \label{eq:vqr1-predicted-probability}
\end{equation}
Compared with the continuous probability in \cref{eq:thurstone}, the official VQ-R1 implementation~\cite{Visualquality-r1} discretizes the MOS relation into three outcomes, breaking the continuity of the original Thurstone formulation, and converts it into a hard label:
\begin{equation}
    y_{ij}=
    \begin{cases}
    1, & \mu_i>\mu_j,\\
    0.5, & \mu_i=\mu_j,\\
    0, & \mu_i<\mu_j.
    \end{cases}
    \label{eq:vqr1-hard-label}
\end{equation}
With the hard label in \cref{eq:vqr1-hard-label}, the fidelity reward is
\begin{equation}
    r_{\mathrm{VQ\text{-}R1},ij}^{(k)}
    =
    \sqrt{p_{ij}^{(k)}y_{ij}}
    +
    \sqrt{(1-p_{ij}^{(k)})(1-y_{ij})}.
    \label{eq:vqr1-fidelity-reward}
\end{equation}
The uncertainty-normalized MR-IQA variant uses human opinion variance through $\tau_{ij}^{\mathrm{unc}}=\sqrt{\sigma_i^2+\sigma_j^2}$, whereas the raw variant sets $\tau_{ij}=1$ and the VQ-R1 baseline~\cite{Visualquality-r1} uses model sampling variance in \cref{eq:vqr1-predicted-probability}.

\subsection{Is Margin Learning Metric Cheating?}
\label{app:metric-optimization-discussion}

Margin learning is not direct PLCC optimization. Quality margin is a relational variable induced by human opinion scores, while PLCC is an evaluation metric. Their connection arises because both remove global offsets and focus on relative variation. During training, MR-IQA uses sampled pairwise errors in main-paper Eq.~(19), not PLCC or the cosine identity in main-paper Eq.~(16), keeping the feedback local and avoiding group-dependent metric rewards.

\clearpage

\twocolumn[{
\section{Qualitative Case Study}
\label{app:case-study}

\vspace{0.4em}
\begin{center}
    \includegraphics[width=0.88\textwidth]{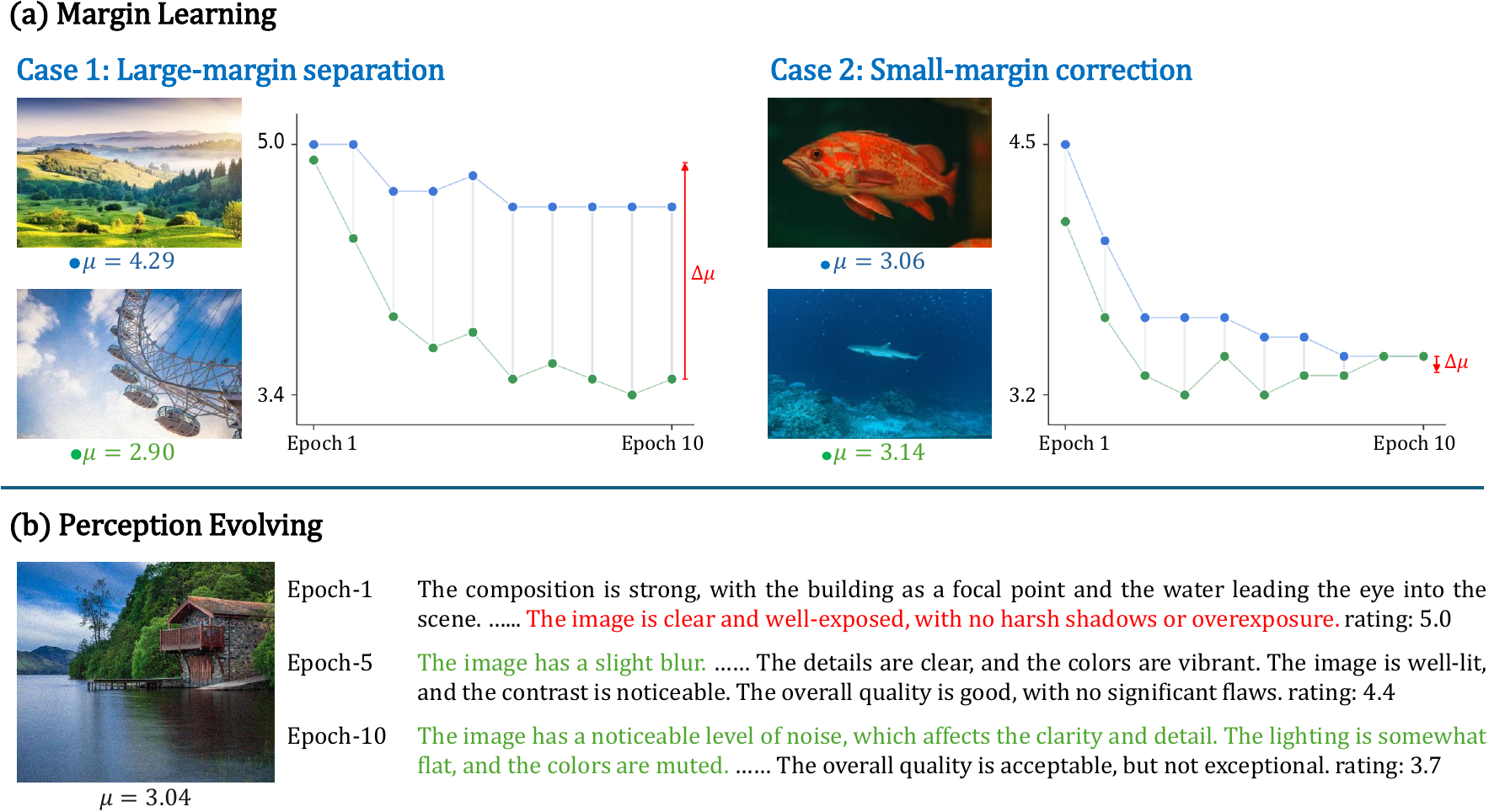}
    \captionof{figure}{\textbf{Qualitative case study of margin learning.} (a) The upper part shows two complementary margin behaviors on validation pairs from KonIQ~\cite{koniq} and KADID-10k~\cite{kadid}: MR-IQA closes an initially overestimated gap for similar-quality images and separates an initially underestimated gap for images with clearer quality differences. (b) The lower part shows the model's gradually increasing perception ability during training, where textual rationales become more sensitive to visible quality degradations.}
    \label{fig:app-case-study}
\end{center}
\vspace{0.4em}
}]

\end{document}